\documentclass{article}

\usepackage[english]{babel}

\usepackage[letterpaper,top=2cm,bottom=2cm,left=3cm,right=3cm,marginparwidth=1.75cm]{geometry}

\usepackage{amsmath}
\usepackage{graphicx}
\usepackage[colorlinks=true, allcolors=blue]{hyperref}

\title{Turtle Score - Similarity Based Developer Analyzer}
\author{
  Sanjjushri Varshini\\
  \texttt{sanjjushrivarshini@gmail.com}
  \and
  Ponshriharini V\\
  \texttt{ponshriharini@gmail.com}
  \and
  Santhosh Kannan\\
  \texttt{mail2santhoshkannan@gmail.com}
  \and
  Snekha Suresh\\
  \texttt{snekhasuresh2777@gmail.com}
  \and 
  Harshavardhan Ramesh\\
  \texttt{harsha77707@gmail.com}
  \and
  Rohith Mahadevan\\
  \texttt{rohithmahadev30@gmail.com}
  \and
  Raja CSP Raman\\
  \texttt{raja@tactii.com}
}

\begin{document}
\maketitle

\begin{abstract}
In day-to-day life, a highly demanding task for IT companies is to find the right candidates who fit the companies’ culture. This research aims to comprehend, analyze and automatically produce convincing outcomes to find a candidate who perfectly fits right in the company. Data is examined and collected for each employee who works in the IT domain focusing on their performance measure. This is done based on various different categories which bring versatility and a wide view of focus. To this data, learner analysis is done using machine learning algorithms to obtain learner similarity and developer similarity in order to recruit people with identical working patterns. It’s been proven that the efficiency and capability of a particular worker go higher when working with a person of a similar personality. Therefore this will serve as a useful tool for recruiters who aim to recruit people with high productivity. This is to say that the model designed will render the best outcome possible with high accuracy and an immaculate recommendation score.

\end{abstract}


\section{Introduction}
Companies hire candidates with the traditional methods, new techniques that each specific organization follows. But in most of the scenarios, there is an issue of hiring candidates who do not fit the culture of the specified organization. Only a third of companies report that they monitor whether their hiring leads to a satisfactory result. But there is a greater percentage of companies who do not follow up on their employees regarding their skills. 

Moreover, it has become difficult to assess the quality of the candidate and whether they would fit right in with the company culture. A lot of variables and factors come into consideration regarding the quality of a candidate the company wants to hire and these vary from different companies concerning different job positions. Predominantly, the quality of a candidate depends upon different factors such as the candidate's perks, competency, CCI (Continuous Capability Improvement), participatory culture, and so on. Also, the hires are not only for entry-level jobs but across most levels. As the job level increases, the requirements for the candidates the company demands also increase respectively. This has brought in a new problem of ensuring the level of a candidate is on par with the already employed. Also, there is a vast set of competition between the candidates solely over the chances of going to a higher rank in their company. In these types of scenarios, there comes a comparison metric between the preceding candidates who are trying to move for a higher rank. 

To address these issues, this research provides a new metric system for the companies to help in identifying the right set of candidates for employability within their company. Data collected from various candidates is analyzed and a recommendation system based on similarity is created. For this, parameters like turtle score - a score based on the learning analytics score, GitHub score, and error archive score are used. Kaggle's score is also taken into account for the machine learning candidates. Kaggle and Github are very popular among most of the leading companies across the globe and these companies take those scores into account as one of the candidate’s requirements. The reason behind this is the scores that each candidate possesses represent their technical skills over the set of domains they have worked on. These scores act as a good metric to evaluate a candidate and compare them with candidates with similar scores. 

\section{Literature Survey}
\cite{https://doi.org/10.1002/hrm.20119} Employability is a major factor in the sustainable competitiveness within the company as well for the personal growth of the candidate. As a result, a question arises in the mind of everyone - Is the candidate chosen the best option from the pool of choice? Companies have a hard time answering the said question due to a lack of monitoring and bias in the hiring process itself. \cite{2022arXiv220201661M} Implicit bias towards characteristics like race, gender, and sexual orientation has always resulted in persistent inequality and reduces the utility of the employer. So, is there any process that is free from the said problems that can be both reliable and efficient? Various machine learning models are analyzed to find an answer.

\cite{8809154}In the modern competitive world, employers are looking into machine learning models to help them fill their spot of vacancy.\cite{2021arXiv211212463B} One among the models and the most prominent ones uses recommendation systems. A recommendation system is a system that uses information filtering to predict the preferences of a user.\cite{2022arXiv220402338H} A recommendation system does so by using the previous history of the user or by finding similarities with other users. It helps to create dynamic recommendations personalized to each user. Collaborative filtering is the widely used method for creating a recommendation system.\cite{2022arXiv220404633H} Let us consider two users A and B. If A and B are interested in the same product, then their preferences for other products may align more compared to any random user. In the modern data-driven world, recommendation systems play a huge role in supporting the process of decision-making.\cite{2022arXiv220211812P}An organization needs to hire candidates without any discrimination based on gender, color, and many other factors. To overcome this using an ML model will effectively solve this crisis and decrease the bias that human beings cause. One of the major reasons to use the ML model in the hiring methodology is that the model can be acclimated over time to adjust to changing priorities. \cite{Kappus_2021} The search for similar-minded people has been over many decades but the application of Machine Learning makes this job simpler and better. The hashtag-based clustering method with a combination of several rounds of the k-means algorithm and the DBSCAN algorithm is used to form clusters of data that belong to like-minded people. Chaotic data of millions of people can be filtered out easily and working partners for each worker can be assigned by the hirer thus increasing the efficiency of both the workers.

\cite{2017arXiv170709751V} Recruiters in the information technology domain have met the problem of finding appropriate candidates by their skills. In the resume, the candidate may describe one skill in different ways or skills could be replaced by others. The recruiters may not have the domain knowledge to know if one’s skills are fit or not, so they can only find ones with matched skills.

\cite{2020arXiv200612665B} Academic credentials include but are not restricted to diplomas, degrees, certificates, and certifications, which act as a way to attest completion of training or education undertaken by the student. Broadly speaking, these credentials may also attest to the successful completion of any test or exam. Ultimately, they serve as a model of independent validation of the said individual’s possession of the knowledge, skill, and ability needed to carry out a particular task or activity.

The most competent skills that are meant to be present among the candidates that are proclaimed by larger companies are based on Kaggle, Github score, etc., Information regarding one’s Kaggle or Github helps the company to identify the level of skill-set of the candidate. \cite{2021arXiv210711929H} Kaggle is a platform service for data analysis competition that was established in 2010. Corporations and data scientists provide the data and challenges, and users propose such things as analysis models.\cite{artiv0122021} Insights such as the quality of contributions made by a developer on GitHub could provide an initial understanding of a potential candidate’s coding capabilities. Considering the rising popularity of social recruiting and the kind of potential GitHub platform offers, GitHub can be leveraged for software developer recruitment.
\section{Existing Model}
\subsection{Avrio}
Avrio on its own has its own set of hiring processes. At first, they basically use a messenger bot in their hiring process to communicate with their candidates. This is sort of their pre-screening process for their interview. They generate a complete profile from their profile analysis of those candidates who submitted their resumes. Avrio has its own Fitscore for each candidate to compare them based on various factors and the generated profile will be run through every job offering database to see if there are any matches. Also, they have their very own virtual assistant called Rio which is responsible for the candidates' personal profile as it is programmed to record and present the analysis of the candidate by asking different sets of questions that will reflect their skills, capability, and interest set for the available job.
\subsection{Ideal}
Somen and Shaun are the co-founders of Ideal. The idea of this started off from their own personal experience as they were struggling to make data-backed hiring decisions in their decisions. The idea of them having a thought that there might be a better idea never left off and now here they are with their solution. They claim to have their own virtual assistant which directly integrates with the existing clients' applications. The assistant also trains itself from their past hiring decisions from millions of candidates.
They adapt to two main processes. At glance, the first part of the process is Resume screening which is by default performed manually. They don't specifically use any standardized criteria for every resume which is why this system exhibits inherent biasing. For an ideal candidate, the virtual AI quickly identifies the required elements using pattern recognition. The second part is Candidate sourcing. The virtual assistant of Ideal uses client credentials to automatically connect to various third-party candidate websites. The required parameters will be made by the clients so that the algorithm consistently searches those third-party websites for any matching for the candidates.

\subsection{Entelo}
Entelo is one such example that revolves around a talent sourcing software platform using AI to help companies identify quality candidates. The algorithm that the company claims is something that is capable of doing some extreme tasks in a short period of time. To achieve this factor, Entelo has identified over 70 predictive variables which are further used to analyze the data of different candidate profiles. One among the examples of the variables is something like the Linkedin profile of the candidate. The company specifically targets individuals who are not actively seeking employment but still are open to exploring new opportunities.

\section{Distance Similarity Used}
\subsection{Chebyshev}
The Chebyshev distance, also called the Tchebyshev distance, is referred to as the maximum absolute distance. It examines the absolute value of the distance between any two points on the axes. The Chebyshev Distance is mainly used in the game of chess and used in warehouse logistics. The Chebyshev distance is given by the mathematical equation:
\[d(p,q) = max(\left|q_x-p_x\right|,\left|q_y-p_y\right|)\]
\subsection{Canberra}
The Canberra Distance is the measure of any two points of a set in a vector space. The selected pair consists of non-negative real numbers. The Canberra distance is used for comparing ranked lists and used in computer systems to check for any malicious activities or any violation of policy. The mathematical expansion is given as:
\[d(p,q) = \frac{\left|q_x-p_x\right|}{\left|q_x\right|+\left|p_x\right|} + \frac{\left|q_y-p_y\right|}{\left|q_y\right|+\left|p_y\right|}\]
\subsection{Euclidean}
Euclidean distance is the length of the line segment joining two points in an Euclidean space. If the Cartesian coordinates of the points are known, the Euclidean distance can be calculated using the Pythagorean theorem. So it is sometimes called the Pythagorean distance.
The Euclidean distance is symmetric and it is positive for any two distinct points and zero for any point and itself.
\[d(p,q) = \sqrt{(q_x-p_x)^2+(q_y-p_y)^2}\]
\subsection{Manhattan}
Manhattan distance is the sum of the projections of the line segment joining two points onto the coordinate axes in an n-dimensional vector space. In simple words, it is the sum of the absolute difference between all dimensions of two points
It is extensively used in a vast array of fields from regression analysis to frequency distribution.
\[d(p,q) = \left|q_x-p_x\right|+\left|q_y-p_y\right|\]
\subsection{Minkowski}
The Minkowski distance can be considered as a generalization of the Euclidean distance and the Manhattan distance. Minkowski distance of order p can be found using the formula
\[d(p,q) = (\sum_{i=1}^{n}|p_{i} - q_{i}|^{p})^{\frac{1}{p}}\]

Where p is an integer. For p=1 and p=2, the Minkowski distance is reduced to the Manhattan distance and Euclidean distance respectively. In this paper, the Minkowski distance of order 3 has been implemented.

\subsection{Bray-Curtis}
Bray-Curtis distance is used to find how similar the given two candidates are. It is calculated as follows:
\[d(p,q) = \sum|u_{i} - v_{i}|/\sum|u_{i}+v_{i}|\]

The differences between the respective values of the candidates are added together and taken as the numerator. The summation of the sum of respective values is taken as the denominator. The result provides the Bray-Curtis distance. If the distance is close to 0, then it means that the given two candidates are closely related.
\subsection{Cosine}
Cosine similarity and Cosine distance are inversely related to each other. As the distance decreases, similarity increases. So, the value of Cosine distance should be closer to 0 to get a highly similar result. Cosine similarity projects the given two vectors on a multi-dimensional plane and calculates the cosine angle between them. 
\[d(p,q) = cos(\theta)= \frac{\textbf{A}.\textbf{B}} {||\textbf{A}||\,||\textbf{B}||}= \frac{ \sum_{i=1}^{n} A_{i}B_{i} }{\sqrt{\sum_{i=1}^{n} A_{i}^{2} }\sqrt{ \sum_{i=1}^{n} B_{i}^{2} }}\]

Here, A and B are the points projected on the multi-dimensional plane. The cosine angle between A and B is calculated to find the cosine similarity.  
Cosine distance = 1 - Cosine similarity 
If the distance is close to 0, then it means that the given two candidates are highly similar to each other.
\section{Proposed Method}
To overcome this problem, the model tries to find the similarity between candidates. In this paper, a new standardization metric is introduced for measuring a candidate’s potential skills on the technical scale for the recruitment process from a company’s point of view. This metric is termed as “ Turtle Score“.
\subsection{Turtle-Score}
A score based on the learning analytics score, GitHub score, Puzzle meter, Job Shadowing meter, and error archive score is the metric that is used to measure one’s potential skills in the technical scale for the recruitment process. Various Machine Learning schemes are integrated to make the process much more subtle. These scores each help to analyze the characteristics of a candidate in specific problems.
\subsection{Github Turtle Score}
Github plays an important role in every developer’s career. Being a code-hosting platform, Github can decide the experience and coding knowledge of a person. A person can be evaluated just by looking at the contributions made by them and the repositories along with the projects they had worked on. Github score is the sum total of all these outcomes and it mainly focuses on the number of commits that a particular person had contributed to all the repositories over a span of time. This shows the consistency and the working nature of a person within a few seconds thereby saving time.
\subsection{Learning Analytics Turtle Score}
LA score is derived from Learning Analytics which is based on the collection and analysis of different websites and articles that the candidate had come across while working. Learning Analytics is an extension that helps in collecting and storing articles, research papers, websites, etc., which seems to be helpful for a person to solve an error or to complete the coding process. This score can help in determining the interests and skill set of a candidate in a particular area of knowledge by analyzing all the data collected by the individual using the Learning Analytics extension.
\subsection{Kaggle Score}
Kaggle is a well-known website for Machine Learning candidates which offers datasets and Jupyter Notebooks to publish and run ML models. Kaggle is the best option if one is going for machine learning-based tasks. Thus, the Kaggle score would reveal the Machine Learning proficiency of a candidate and can help determine the right candidate for the job accordingly.
\subsection{Error Archive Score}
Errors occupy an important position in the career of a person as it is a part of the learning process and it helps change the perspective of viewing a particular problem. This is why it is essential to keep a note of all the errors and the solution used to overcome them. This score can help in analyzing the problem-solving ability and the technical skill to overcome that error.
\subsection{Job Shadowing Meter}
Job shadowing meters are derived based on how interactive and learnable a particular candidate is during their job shadowing period. A candidate is asked to observe someone else working on a particular technology and gain knowledge from them. During this period, if the candidate is capable of understanding what exactly is going on and if they are able to work on the said technology without any guidance, then they get a high score. This is used to see where a person stands on their learning and observation scale.
\subsection{Puzzle Meter}
Puzzle meters are an exclusive tool to identify candidates with an urge to solve problems given to them no matter how long it takes or how hard the task is. A candidate can be assigned with tasks from different areas of expertise thereby the Manager can assess them by considering the time taken to solve the particular problem and the interest of the candidate in that particular task which made them work 
off the clock. This helps in filtering out the candidates with talent in a particular field of interest.
The company can assign each and every candidate with a custom-made score according to their performance in the tasks assigned. Thus, deserved candidates can be chosen for the jobs accordingly.

Therefore by combining all these scores, a person would be able to determine the total skill-set of a candidate and also find similar candidates with the same talents. The similarity between any two candidates is found by comparing the above parameters between any two candidates. This similarity found can in turn be used to find the potential job position any given candidate might be able to get.
\section{Technology Used}
\subsection{Flask}
This web framework is used to display the result in a web interface. The application must be brought in this interface form in order to deploy it using docker. Flask framework in this model takes in a user’s details as input and provides with similar candidate list as output along with its similarity percentile. It also predicts and displays the job position which might be suitable for the candidate based on the given candidate data. This comparison takes place in the background using the code present in a python file.
\subsection{Docker}
A simple docker deployment is done in order to avoid any errors related to versions. The current machine learning model runs in a particular version with which it is built. When it is used in a different system, it might throw some error depending on the version used in that user’s system. To overcome this problem, a docker image with all the versions specified is created. Since the docker image installs the required versions, the application runs perfectly fine in any system without errors related to its version. 
\section{Working}
\subsection{Data Flow}
\begin{figure}[htp]
\centering
\includegraphics[width=1.1\textwidth]{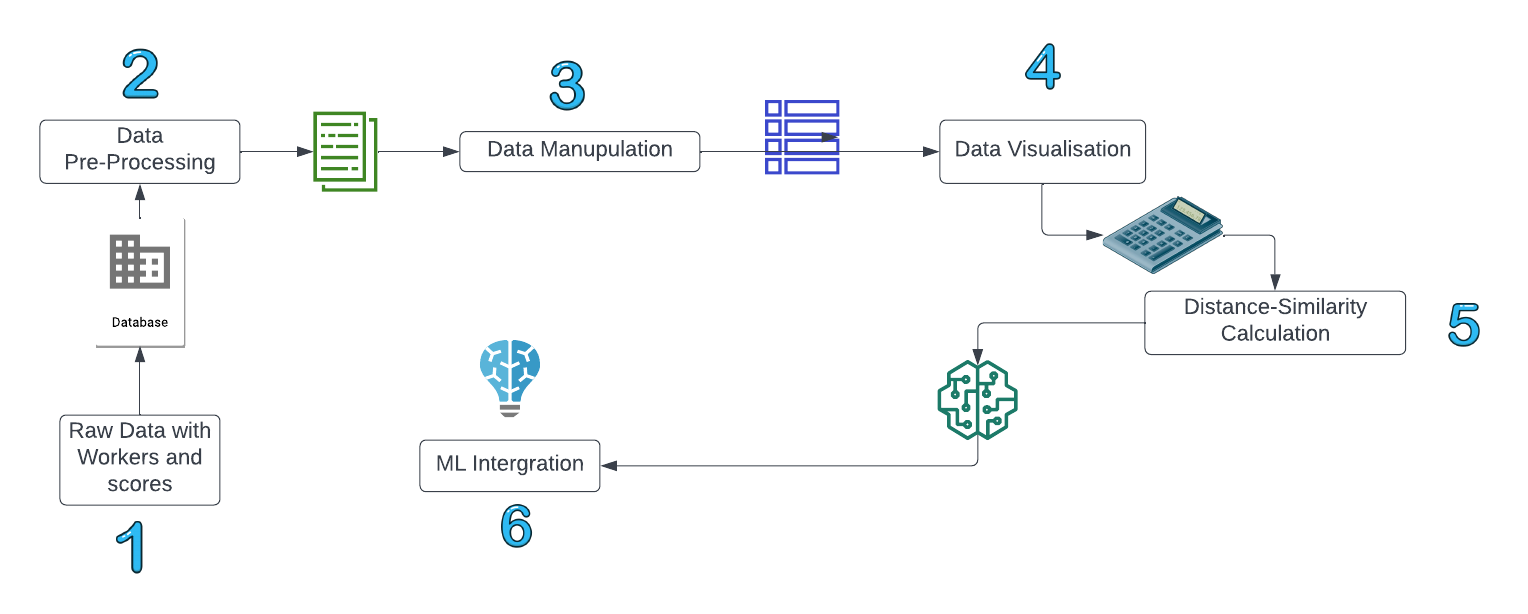}
\caption{\label{Fig 1: Data flow}Data Flow}
\end{figure}
\subsubsection{Raw Data}
Real-time data of each candidate is piled and dumped into the database.
\subsubsection{Data Pre-Processing}
The data would be filtered and would undergo a few processes to remove outliers and to fill the null values.
\subsubsection{Data Manipulation}
The data is visualized for a better understanding of the workflow of each candidate and to compare among the candidates to find similar ones.
\subsubsection{Distance-Similarity Calculation}
The difference among each candidate's data is calculated and an overall similar percentile score is found. 
\subsubsection{ML Integration}
This includes the prediction of similar candidates using the related similar score accordingly using the machine learning tools and the accurate results are produced. 
\subsection{Data Architecture}
\begin{figure}[htp]
\centering
\includegraphics[width=1.0\textwidth]{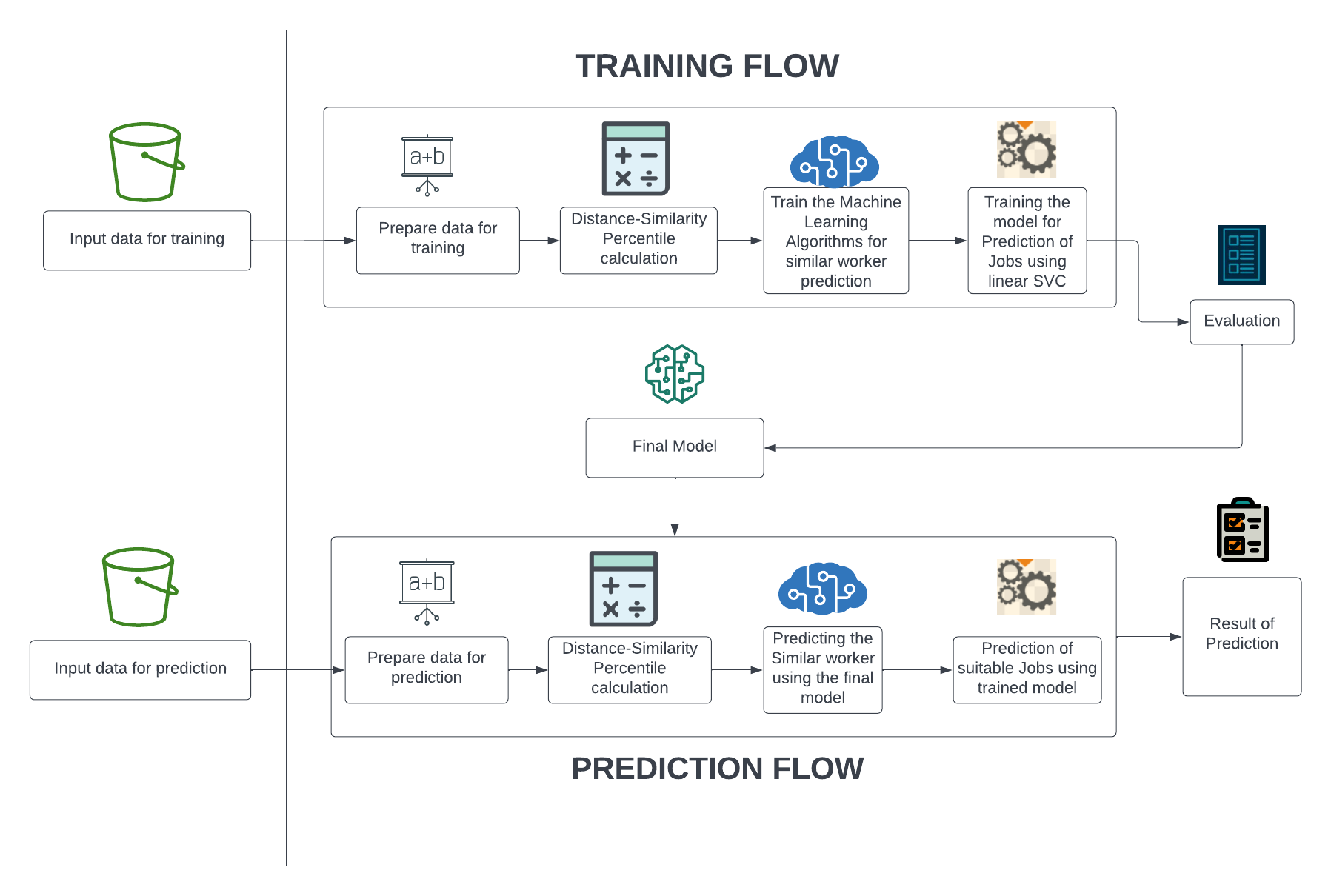}
\caption{\label{Fig 1: Architecture chart}Architecture Chart}
\end{figure}
\subsubsection{Training Flow}
The inputs for the training model will be loaded. The input data set that has been loaded for the training model is the company dataset of all candidates. Now, the data is prepared for training where it will undergo some process to fill null values and remove outliers in the provided data set. Now the prepared data set will be tested and used in different distance similarity measures for calculating the distance-similarity percentile. Now the machine learning algorithm will be trained for some similar worker prediction. Now as the model is further trained for job prediction under linear SVC. The resulting data will be sent for final evaluation and the final model will be presented.
\subsubsection{Prediction Flow}
As the training flow gets completed, the system will now load inputs for the prediction model. The Input data set that has been loaded for the training model is the company dataset of all candidates. Now, the data is prepared for training where the data will undergo some process to fill null values and remove outliers in the provided data set. Now the prepared data set will be tested and used in different distance similarity measures for calculating the distance-similarity percentile. Now the resulting values will be sent for prediction analysis for finding a similar worker using the final model. Now as the model is further trained for similar jobs to the predicted model. The system will now present the result that was produced by the prediction model. 
\section{Result Analysis}
To find the most similar candidate to any given candidate, different models to get varying results are used. Each model has its own ups and downs, so these models here are analyzed and their results are compared to one another. These results are calculated based on the turtle score. Therefore the comparison is done based on this and similarity is then calculated accordingly. It depends on how closely they are related, that is, the smaller the difference between the scores of any two candidates, the closer is the relation between them.
\begin{figure}[htp]
\centering
\includegraphics[width=1.0\textwidth]{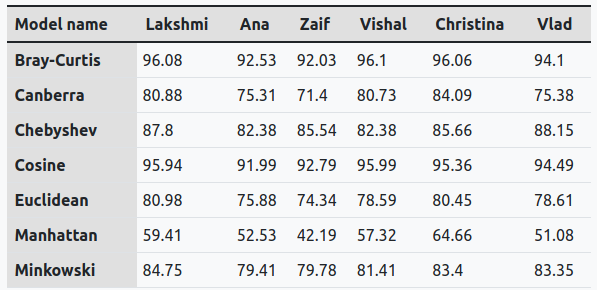}
\caption{\label{Fig 1: Similarity percentile table}Similarity percentile table}
\end{figure}
Fig. 3 indicates the similarity percentile for 6 different people when they are compared to a given candidate. This table shows how the result varies depending on the distance calculation function used. Thus, based on distance and percentile range, models can be ordered as least to most similar candidate finder:
\begin{enumerate}
\item Manhattan
\item Euclidean
\item Minkowski
\item Canberra
\item Chebyshev
\item Bray-Curtis
\item Cosine
\end{enumerate}
Also, the most appropriate job for the candidate is predicted. For this, a classification model is trained on the turtle score of the company employees. Next, it is fed with the candidate’s turtle score to predict the suitable job for the candidate in the company. Several supervised classification models were tested and Linear Support Vector Classifier (SVC) was the best among them with an accuracy of 87

\section{Future Works}
Over this course of Turtle score research, the comparison is done using Distance Similarity measures such as Euclidean, Manhattan, Minkowski, and many more. There are over 9 different Distance similarity measures used in this research and further research can be done with many more similarity measures. Further research can also be done by bringing up many more entities. This might lead to a much closer similarity. Also, the addition of different unique entities makes the turtle score more compatible and customizable as organizations from different sectors can use it as well.

Data can also be derived from various candidates belonging to various age groups. This will provide the dataset with diversity which in turn will improve the application range of the model.

\section{Conclusion}
The primary ideology behind this applied research is that the process of recruitment is made to be efficient in order to hire candidates like whom exactly the company wishes to have. This can be achieved by the innovative algorithm methodology this paper deals with. This can even study the student and help them to identify their own standard and can sustain in IT industries. This will make the job of the recruiting strategy more manageable in such a way that companies can employ the individuals they hope to. Various algorithms and testing were implemented to deliver the best outcomes.

One of the biggest pros of Turtle score is that it is compatible with any environment. This turtle score can also be implemented in more than one set of organizations such as 
\begin{itemize}
   \item It can be modified in such a way that the schools can track the progress of their students over the years,
   \item Helps medical organizations to understand the patient's health in terms of recovering from any disease or any syndrome and many more.
\end{itemize}
It can also be used to find what kind of jobs are compatible for the given candidate. This can be done by getting a similar candidate and checking their current profession or job post. Therefore, this model can be used to predict a person’s future job using their current data. This makes it possible to apply the turtle score model to any college-going or school student and predict their future job. It also helps to provide a progress report which tells them where to improve in order to land a particular job post. This in turn encourages students to stay on their feet and improve.

Finding similar candidates not only makes it possible to recruit people based on the requirements but also provides more data about a person as similar data is stored in the dataset. It’ll serve as an effective tool in identifying potential amongst people in the IT sector or it can even spread out to other sectors.
\section{Acknowledgement}

We would like to show our gratitude to Featureprenuer for sharing their valuable data for the development of this application.

\bibliographystyle{plain}
\bibliography{reference.bib}

\begin{thebibliography}{10}

\bibitem{2020arXiv200612665B}
Chaitanya {Bapat}.
\newblock {Blockchain for Academic Credentials}.
\newblock {\em arXiv e-prints}, page arXiv:2006.12665, June 2020.

\bibitem{2021arXiv211212463B}
Hrisav {Bhowmick}, Ananda {Chatterjee}, and Jaydip {Sen}.
\newblock {Comprehensive Movie Recommendation System}.
\newblock {\em arXiv e-prints}, page arXiv:2112.12463, December 2021.

\bibitem{2021arXiv210711929H}
Teruaki {Hayashi}, Takumi {Shimizu}, and Yoshiaki {Fukami}.
\newblock {Collaborative Problem Solving on a Data Platform Kaggle}.
\newblock {\em arXiv e-prints}, page arXiv:2107.11929, July 2021.

\bibitem{2022arXiv220404633H}
Heidy {Hazem}, Ahmed {Awad}, and Ahmed {Hassan}.
\newblock {A Distributed Real-Time Recommender System for Big Data Streams}.
\newblock {\em arXiv e-prints}, page arXiv:2204.04633, April 2022.

\bibitem{https://doi.org/10.1002/hrm.20119}
Claudia M. Van~Der Heijde and Beatrice I. J.~M. Van Der~Heijden.
\newblock A competence-based and multidimensional operationalization and
  measurement of employability.
\newblock {\em Human Resource Management}, 45(3):449--476, 2006.

\bibitem{2022arXiv220402338H}
Jun {Hu}, Shengsheng {Qian}, Quan {Fang}, and Changsheng {Xu}.
\newblock {MGDCF: Distance Learning via Markov Graph Diffusion for Neural
  Collaborative Filtering}.
\newblock {\em arXiv e-prints}, page arXiv:2204.02338, April 2022.

\bibitem{Kappus_2021}
Philipp Kappus and Paul Gro{\ss}.
\newblock Finding clusters of similar-minded people on twitter regarding the
  covid-19 pandemic.
\newblock In {\em Machine Learning Techniques and Data Science}. Academy and
  Industry Research Collaboration Center ({AIRCC}), nov 2021.

\bibitem{8809154}
Ali~A. Mahmoud, Tahani AL~Shawabkeh, Walid~A. Salameh, and Ibrahim Al~Amro.
\newblock Performance predicting in hiring process and performance appraisals
  using machine learning.
\newblock In {\em 2019 10th International Conference on Information and
  Communication Systems (ICICS)}, pages 110--115, 2019.

\bibitem{2022arXiv220201661M}
Anay {Mehrotra}, Bary S.~R. {Pradelski}, and Nisheeth~K. {Vishnoi}.
\newblock {Selection in the Presence of Implicit Bias: The Advantage of
  Intersectional Constraints}.
\newblock {\em arXiv e-prints}, page arXiv:2202.01661, February 2022.

\bibitem{artiv0122021}
Hauff~C. Nagaram~A.R.
\newblock {Using Github Profiles in Software Developer Recruitment and Hiring}.
\newblock {\em TUDelft}, August 2021.

\bibitem{2022arXiv220211812P}
Andi {Peng}, Besmira {Nushi}, Emre {Kiciman}, Kori {Inkpen}, and Ece {Kamar}.
\newblock {Investigations of Performance and Bias in Human-AI Teamwork in
  Hiring}.
\newblock {\em arXiv e-prints}, page arXiv:2202.11812, February 2022.

\bibitem{2017arXiv170709751V}
Le~{Van-Duyet}, Vo~Minh {Quan}, and Dang {Quang An}.
\newblock {Skill2vec: Machine Learning Approach for Determining the Relevant
  Skills from Job Description}.
\newblock {\em arXiv e-prints}, page arXiv:1707.09751, July 2017.

\end{thebibliography}

\end{document}